\documentclass{article}
\usepackage{ijcai26}

\usepackage{amssymb}
\usepackage{bm}
\usepackage{times}
\usepackage{soul}
\usepackage{url}
\usepackage[hidelinks]{hyperref}
\usepackage[utf8]{inputenc}
\usepackage[small]{caption}
\usepackage{graphicx}
\usepackage{amsmath}
\usepackage{amsthm}
\usepackage{booktabs}
\usepackage{algorithm}
\usepackage{algorithmic}
\usepackage[switch]{lineno}
\usepackage{makecell}
\usepackage{multirow}
\usepackage{makecell}
\renewcommand{\arraystretch}{0.5}
\setcellgapes{1pt} 
\makegapedcells  

\title{Mitigating Entity Type Confusion in Cross-Domain NER via Multidimensional Quantification and Reasoning Enhancement}

\author{
Jingyu Wang\thanks{Corresponding authors.} 
\and
Shijie Wu
\and
Fusheng Jin$^{*}$\\
\affiliations
Beijing Institute of Technology\\
\emails
jyw.bit.2003@gmail.com,
wsj1595@bit.edu.cn,
jfs21cn@bit.edu.cn  
}

\begin{document}

\maketitle

\begin{abstract}
Cross-domain Named Entity Recognition (CD-NER) aims to transfer the rich knowledge in the source domain to the target domain.
Recent studies adopting decomposition or generation paradigms have achieved significant performance improvements, demonstrating high accuracy in entity span detection. However, during entity type classification, models severely suffer from entity type confusion, the erroneous tendency that models classify entities of one type in the text as another similar but incorrect type.
To address this issue, we first propose a Multidimensional Confusion Quantification Model (MCQM) that quantifies a model's confusion extent between entity types from three dimensions: source-target hierarchy analysis, semantic similarity analysis, and explicit data evaluation. Moreover, we propose the Progressive Bidirectional Reasoning Chain (PBRC). PBRC leverages the source-target hierarchy and confusion analysis from the MCQM to prompt the LLM to generate two-stage reasoning information. The two-stage reasoning information is utilized to augment the knowledge of the model, significantly mitigating entity type confusion and improving the model's generalization performance.
Experimental results demonstrate that our method achieves new state-of-the-art results on all domains of the CrossNER dataset.
\end{abstract}

\section{Introduction}
Named Entity Recognition (NER) is a core task in information extraction, which aims to identify specific entities in texts that belong to predefined types, such as person, location, and organization~\cite{1,26,55}. NER plays a critical role in various tasks, including information retrieval~\cite{6,7,9}, knowledge graphs~\cite{3,5,61}, and recommendation~\cite{10,60}. With large-scale annotated data, neural network-based methods have performed remarkably in traditional NER~\cite{16,37,59}. However, in cross-domain NER, several challenges persist, especially in low-resource scenarios where model performance experiences a significant decline.
\begin{figure}
    \centering
    \includegraphics[width=1\linewidth]{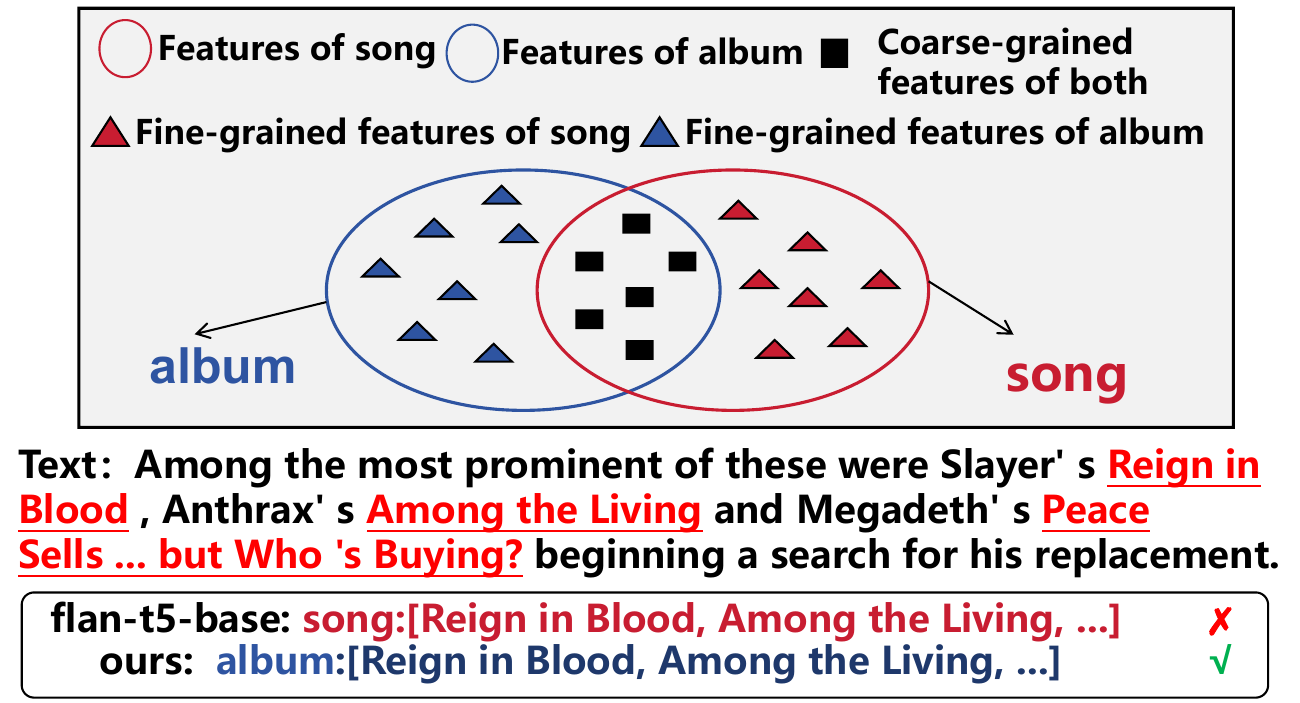}
    \caption{{\fontseries{m}\textit{song}} and {\fontseries{m}\textit{album}} share similar contextual features and entity forms, which can lead to confusion.}
    \label{fig:1}
\end{figure}

Recent studies~\cite{11,36,27} decompose cross-domain NER into subtasks to capture transferable patterns across domains. These methods focus on identifying shared feature patterns between domains through specialized components to strengthen source-domain knowledge transfer. 
Other approaches \cite{13,53,48} employ generative pre-trained language models, transforming NER into a generation task through the use of task-specific input prompts. These methods can effectively utilize large-scale knowledge from pretraining and exhibit high flexibility. Although both paradigms have demonstrated effectiveness in improving cross-domain NER performance, they remain susceptible to entity type confusion, leading to significant performance degradation in entity type classification.

We observe that cross-domain NER currently faces two major challenges. (1) Granularity differences between coarse-grained entity types in the source domain and fine-grained entity types in the target domain. Entity types in the source domain are coarse-grained, such as PER and LOC, whereas entity types in the target domain are more specific and domain-specialized, such as scientist and country. Compared to coarse-grained types, distinguishing fine-grained types often requires more detailed and nuanced features. Relying solely on the coarse-grained features learned from the source domain makes it difficult for the model to accurately classify fine-grained types in the target domain, such as song and album (as shown in Figure~\ref{fig:1}). (2) The lack of contextual information in the text further limits the model’s ability to classify fine-grained entity types accurately. In the target domain, some samples lack the necessary contextual cues, making it difficult for the model to effectively learn and capture the required fine-grained features for type classification. As shown in Figure~\ref{fig:1}, we selected a sample from CrossNER~\cite{24}, where the song and the album are two entity types that are difficult to distinguish. Entities of both types share similar forms and contextual environments. Due to insufficient information in the text, the model misclassifies these entities as the song. The lack of adequate contextual information prevents the model from effectively learning the feature differences between similar fine-grained entity types, which reduces the precision of type classification.

To address the challenges in cross-domain NER, we propose the Multidimensional Confusion Quantification Model (MCQM) and the Progressive Bidirectional Reasoning Chain (PBRC). (1) We propose the MCQM that quantifies the model's confusion extent from three dimensions: source-target hierarchy analysis, semantic similarity analysis, and explicit data evaluation. MCQM enables in-depth analysis and quantification of the model's confusion extent between fine-grained entity types in the target domain, providing a foundation for addressing the entity type confusion. (2) Furthermore, we innovatively leverage the source-target hierarchy, along with the confusion analysis from the MCQM, to propose the PBRC, which employs a two-stage reasoning strategy. Specifically, PBRC first instructs the LLM to perform initial reasoning based on contextual information using coarse-grained entity types from the source domain. Then, it instructs the LLM to perform bidirectional reasoning using fine-grained entity types from the target domain and confusion analysis from the MCQM. Finally, the LLM generates two-stage reasoning information for knowledge augmentation.
The two-stage reasoning strengthens the fine-grained features and alleviates the granularity gap between the source and target domains, while external knowledge provided by the LLM solves the issue of insufficient contextual information. Our method significantly mitigates entity type confusion. Experimental results demonstrate that our method effectively improves the model's generalization ability, with the average F1 score increasing by more than 10.00\%. 
In summary, our contributions are as follows:
\begin{itemize}
\item We systematically analyze the entity type confusion in cross-domain NER. Then, we propose the MCQM to quantify the model's confusion extent effectively. 
\item We propose the PBRC based on the source-target hierarchy and confusion analysis from the MCQM, which can leverage the LLM to mitigate entity type confusion in cross-domain NER significantly.
\item Extensive experiments show that our method significantly improves the model's generalization ability, achieving new state-of-the-art performance across all domains of the CrossNER dataset.
\end{itemize}

\section{Related Work}
\textbf{Named Entity Recognition (NER).} Traditional methods rely on handcrafted features or rules combined with machine learning models like CRF~\cite{32} for entity recognition, which are labor-intensive and lack adaptability to complex contexts. With the development of deep learning, neural network-based methods, such as LSTM~\cite{33} and transformer~\cite{34} models, have become mainstream. Models~\cite{17,18,35} based on these architectures capture sequential relationships and contextual dependencies, offering improved performance and flexibility. However, existing methods still face challenges in cross-domain NER due to the variations across different domains.

\begin{figure*}
    \centering
    \includegraphics[width=1\linewidth]{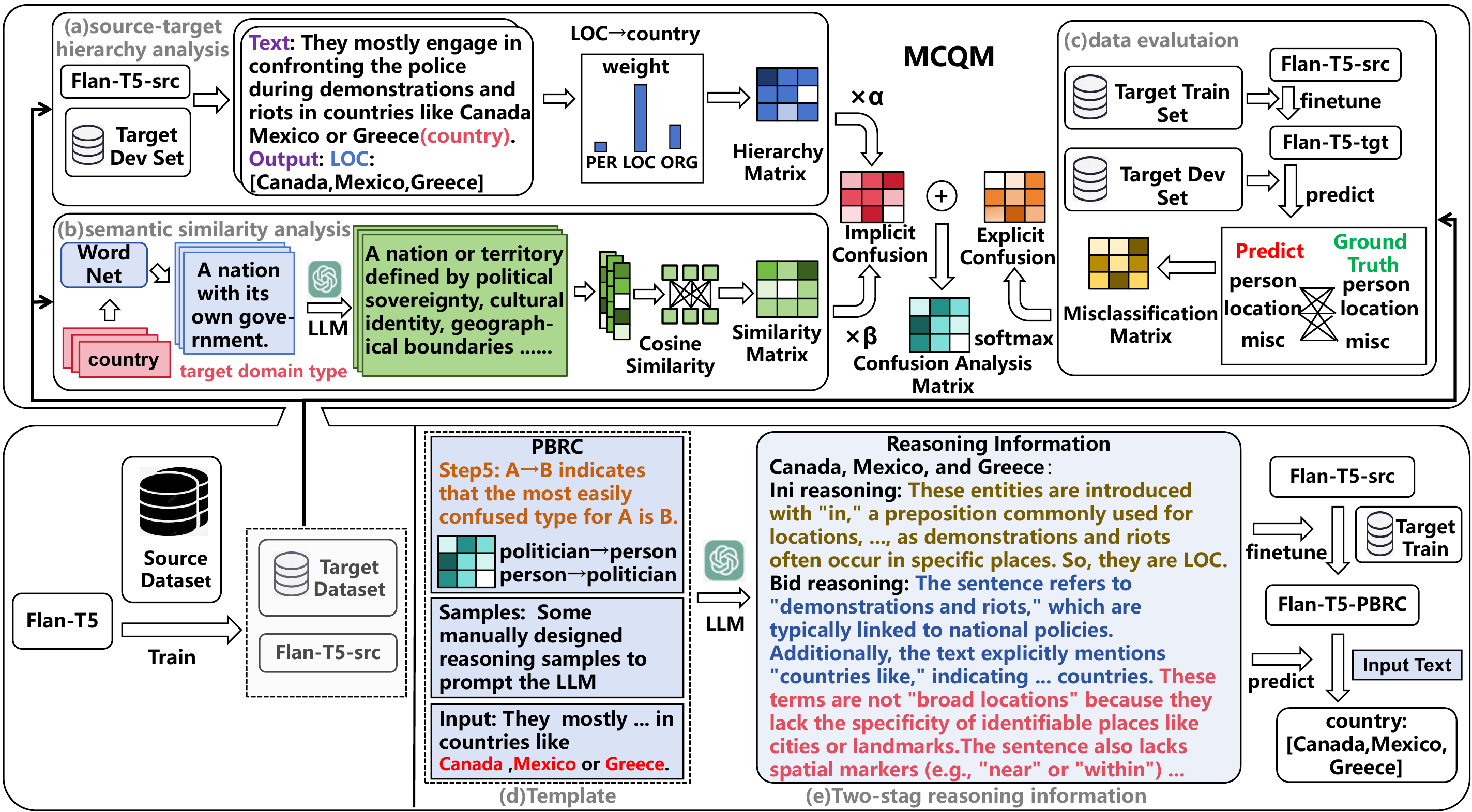}
    \caption{Overview of our proposed method, including MCQM and PBRC.
    }
    \label{fig:2}
\end{figure*}

\textbf{Cross-Domain NER.} 
Current studies can be broadly categorized into two paradigms: decomposition-based and generation-based. Decomposition-based methods~\cite{11,27,36,49} reformulate cross-domain NER into multiple subtasks. These methods improve knowledge transfer by learning shared cross-domain patterns through specialized subtask modules. 
Generation-based methods~\cite{12,13,53,48} transform the NER into a generation task. These methods employ prompts containing task descriptions and auxiliary information, achieving robust performance through instruction fine-tuning. Although both paradigms effectively improve cross-domain performance, particularly in entity span detection, they still suffer from significant accuracy degradation in entity type classification.

\textbf{LLMs for NER.} 
With the emergence of LLMs, their reasoning capabilities have been leveraged to improve NER performance. Recent studies~\cite{41,42,43,21} employ prompt learning for NER, where task-specific templates are augmented with demonstration examples and knowledge. LLMs are also used for cross-domain NER. Nandi and Agrawal~\cite{48} perform instruction fine-tuning of LLMs by retrieving and leveraging similar examples. Zhang et al.~\cite{53} utilize the LLM to generate task-oriented knowledge, used to conduct additional task-oriented pre-training of the backbone model for domain adaptation. These methods demonstrate strong few-shot performance. 

Compared to fine-tuning LLMs~\cite{48} and additional pre-training corpora~\cite{53}, our method reduces training time for domain adaptation.

\section{Methodology}
\subsection{Task Description}
\label{sec:3.1}
Given a sentence $X = \{x_1, x_2, \dots, x_N\}$, where $w_i$ represents the $i$-th token in the sentence and $N$ is the sentence length, the goal of NER is to identify all entities $E$ within the sentence and classify each entity into a specific type. The set of entity types is denoted as $L$. Each entity $e_i \in E$ can be represented as $e_i = (y, x_{l:r})$, where $y \in L$ indicates the entity type, and $l$ and $r$ represent the start and end boundary indexes of the entity within the sentence, respectively. In the cross-domain NER, two distinct datasets are considered: the source domain dataset $\mathcal D_{src}$ and the target domain dataset $\mathcal D_{{tgt}}$. The set of entity types in the source domain is denoted as \( L_s = \{s_1, s_2, \dots, s_n\} \), and the set of entity types in the target domain is denoted as \( L_t = \{t_1, t_2, \dots, t_m\} \). The goal is to leverage the knowledge learned from $\mathcal D_{{src}}$ to improve recognition performance on $\mathcal D_{{tgt}}$. 
Specifically, we focus on low-resource scenarios, where the training data in the target domain is significantly smaller than that in the source domain, i.e., $\lvert \mathcal D_{{tgt}} \rvert \ll \lvert \mathcal D_{{src}} \rvert$.

\subsection{Multidimensional Confusion Quantification Model}
\label{sec:3.2}
Entities of the same type share certain common features, referred to as entity type features, and entity types with similar features are more prone to confusion.  
We quantify entity type confusion from three dimensions in Figure~\ref{fig:2}. 
\subsubsection{Source-Target Hierarchy Analysis}
\label{sec:3.2.1}
Entity types in the source domain are conceptually coarse-grained, whereas those in the target domain are fine-grained. In the feature space, the fine-grained entity types in the target domain can be viewed as extensions of the entity types in the source domain. For example, PER in the source domain can be expanded into fine-grained entity types in the target domain, such as politician and scientist. The expanded entity types retain the coarse-grained features in the source domain. Therefore, fine-grained entity types derived from the same entity type in the source domain exhibit similar features, meaning closer in the feature space and more prone to confusion among these entity types.

To quantify the model's confusion arising from the source-target hierarchy between the source and target domains, inspired by~\cite{12}, we propose a reliable quantification method. For each entity in the target domain, its feature vector $ \bm{\mathcal{H}}_e $ can be considered as a combination of coarse-grained features $\bm H_{coarse}$ and fine-grained features $\bm H_{fine}$. 

We only focus on extracting $\bm {\mathcal{H}}_{coarse}$. We utilize the model \( \mathcal M_{src} \), trained on \( \mathcal{D}_{src} \), to extract \( \bm {\mathcal{H}}_{coarse} \) of entities in the target domain and perform predictions based on the source domain type set \( L_s \). We calculate the proportion of entities corresponding to each fine-grained type \(t_i\) in the target domain \( \mathcal{D}_{tgt} \) that are classified as the coarse-grained type \(s_j\) in the source domain \( \mathcal{D}_{src} \):
\begin{equation}
R_H\left(t_i \rightarrow s_j\right)=\frac{P\left(l_{tgt}=t_i \wedge l_{pred}=s_j\right)}{P\left(l_{tgt}=t_i\right)}
\end{equation}
where \(P(l_{{tgt}} = t_i)\) is the proportion of entities in \(\mathcal{D}_{{tgt}}\) that belong to type \(t_i\), \(P(l_{{tgt}} = t_i \land l_{{pred}} = s_j)\) is the proportion of entities in \(\mathcal{D}_{{tgt}}\) that belong to type \(t_i\) and are classified as \(s_j\), and the proportion of entities of type \(t_i\) classified as \(s_j\) is denoted as \({R_H}(t_i \to s_j)\).

We use the source domain entity type \(l_{{src}}\) with the highest proportion as the prefix for the target domain entity type \(l_{{tgt}}\), denoted as \(l_{{src}} \to l_{{tgt}}\). The target domain entity type \(l_{{tgt}}\) is considered a fine-grained extension of the source domain entity type \(l_{{src}}\):
\begin{equation}
prefix\left(t_i\right)=\mathop{\arg\max}\limits_{s_j \in L_s} \left(R_H\left(t_i \rightarrow s_j\right)\right)
\end{equation}
where \({prefix}(t_i)\) is the prefix of \(t_i\), and $L_s$ is the set of entity types in the source domain.

\( {R_H}(t_i \to s_j) \) measures how much of the coarse-grained features of \( s_j \) are contained in \( t_i \). The more entities corresponding to \( t_i \) are classified as \( s_j \), the more coarse-grained features of \( s_j \) are contained in \( t_i \). Therefore, for two fine-grained entity types, \( t_a \) and \( t_b \), if they share the same prefix, it implies that \( t_a \) and \( t_b \) exhibit a significant overlap in their coarse-grained features, which originate from the same source domain entity type. This overlap increases the likelihood of confusion between \( t_a \) and \( t_b \). We quantify the confusion arising from the source-target hierarchy using \({R_H}(t_i \to s_j) \), based on the prefixes of fine-grained entity types:
\begin{equation}
u_{hie}^{\left(t_i, t_j\right)}=\delta_{hie } \cdot R_H \left(t_i \rightarrow prefix\left(t_j\right)\right)
\end{equation}
where \( u_{hie}^{(t_i, t_j)} \) is the extent of hierarchical confusion from the fine-grained entity type \( t_i \) to \( t_j \), and \( \delta_{{hie}} \) is the scaling factor. This method quantifies the model's confusion arising from the source-target hierarchy, serving as an important metric in the MCQM for quantifying entity type confusion.

\subsubsection{Semantic Similarity Analysis}
\label{sec:3.2.2}
The semantics of entity types reflect the common features of their entities. When the model maps the embedding vectors of the semantics of two entity types to closer positions in the semantic space, it considers the entity features of the two types as more similar, which increases the probability of confusing the two types in classification.

Since entity type labels are short words, they cannot sufficiently reflect the features of the entity types. To enrich the semantics of each target domain entity type \( t_{i} \in L_{{t}} \), we utilize the WordNet~\cite{31} to obtain a brief description \( C_s \) of \( t_{{i}} \). Then, $C_s$ is further refined into a final description $C_e$ of $t_{i}$ by leveraging the LLM. Then, \( C_e \) is fed into the encoder of the model, from which the hidden layer feature sequence \( \bm{\mathcal{H}} = [ \bm h_1, \bm h_2, \dots, \bm h_Z ] \in \mathbb{R}^{Z\times d} \) can be extracted:
\begin{equation}
\bm {\mathcal{H}}=\operatorname{Encoder}\left(C_e\right)
\end{equation}
where \( \bm h_i \) denotes the feature vector from the final hidden layer for the \( i \)-th token, \( Z \) is the number of tokens, and \( d \) represents the dimension of the final hidden layer of the encoder.
This study~\cite{56} demonstrates that Mean Pooling outperforms both [CLS] token and Max Pooling strategies in semantic textual similarity tasks. Therefore, we use Mean Pooling to calculate the global feature vector \( \bm {\overline{\mathcal{H}}} \), which better integrates the semantic information of all tokens and provides a more comprehensive representation of the entity type features:
\begin{equation}
\bm {\overline{\mathcal{H}}}=\frac{\sum_{i=1}^Z m_i \cdot \bm h_i}{\sum_{i=1}^Z m_i}
\end{equation}
where \( m_i \) is the mask value of the \(i\)-th token, used to ignore the influence of padding.

In high-dimensional space, the features of entity types are amplified, and the mapping rules for similar features are also similar. Therefore, we calculate the semantic similarity between entity types using \( \bm {\overline{\mathcal{H}}} \). We quantify the confusion arising from the semantic similarity by computing the cosine similarity between the global feature vectors \( \bm {\overline{\mathcal{H}}} \) of entity types:
\begin{equation}
u_{\text{sim}}^{\left(t_i, t_j\right)}=\frac{\bm {\overline{\mathcal{H}}}_i \cdot \bm {\overline{\mathcal{H}}}_j}{\left\|\bm {\overline{\mathcal{H}}}_i\right\| \cdot \left\|\bm {\overline{\mathcal{H}}}_j\right\|}
\end{equation}
where \( u_{sim}^{(t_i,t_j)} \) is the extent of semantic confusion from the fine-grained entity type \( t_i \) to \( t_j \). This method effectively quantifies the model's confusion arising from the semantic similarity between entity types, serving as an important metric in the MCQM for quantifying entity type confusion.

\subsubsection{Fusion of Implicit Confusion Factors}
\label{sec:3.2.3}
The quantification of confusion arising from the source-target hierarchy and the semantic similarity of entity types is considered implicit confusion analysis. We use two fixed mixing ratios, \( \alpha \) and \( \beta \), to balance the weights of these two factors:
\begin{equation}
\begin{split}
u_{iml}^{\left(t_i, t_j\right)} = \alpha \cdot u_{hie}^{\left(t_i, t_j\right)} 
+ \beta \cdot u_{sim}^{\left(t_i, t_j\right)}
\end{split}
\end{equation}

Then, we use the softmax function to model the implicit confusion and define the implicit confusion distribution among fine-grained entity types in the target domain:
\begin{equation}
v_{iml}(t_i \rightarrow t_j) = 
\frac{
  \exp(w_0 \cdot u_{iml}^{(t_i, t_j)})
}{
  \sum\limits_{\substack{t_k \in L_t \\t_k \neq t_i}} \exp(w_0 \cdot u_{iml}^{(t_i, t_k)})
}
\end{equation}
where \( w_0 \in \mathbb{R}^+ \) is a temperature coefficient.

\subsubsection{Explicit Data Evaluation}
\label{sec:3.2.4}
We denote the model obtained by fine-tuning \( \mathcal M_{{src}} \) on the target domain dataset as \( \mathcal M_{{tgt}} \). We use the dev set of the target domain to perform a statistical analysis of the misclassification proportions for all entity types. Evaluating \( \mathcal M_{{tgt}} \) reflects the model's ability to distinguish between different fine-grained entity types and explicitly reveals the model's confusion extent between entity types:
\begin{equation}
R_M^{(t_i , t_j)} = \frac{P(l_{true} = t_i \land l_{pred} = t_j)}{P(l_{true} = t_i)}
\end{equation}
where \( P(l_{{true}} = t_i) \) is the proportion of entities that truly belong to type \( t_i \), 
\( P(l_{{true}} = t_i \land l_{{pred}} = t_j) \) is the proportion of entities that truly belong to type \( t_i \) but are misclassified as \( t_j \), and the proportion of entities of type \( t_i \) misclassified as \( t_j \) is denoted as \( R_M^{(t_i, t_j)} \).

The value of \(R_M^{(t_i, t_j)} \) explicitly reflects the model's learning effectiveness in distinguishing the fine-grained features of \( t_i \) and \( t_j \). 
A larger value indicates that \( t_i \) and \( t_j \) are more prone to confusion. 
We use \(R_M^{(t_i, t_j)} \) as an important metric for confusion analysis in the MCQM. Then, we use the softmax function to model explicit confusion and define the explicit confusion distribution among fine-grained entity types in the target domain:
\begin{equation}
v_{exl}(t_i \rightarrow t_j) = 
\frac{
  \exp ( w_0 \cdot R_M^{(t_i , t_j)})
}{
  \sum\limits_{\substack{t_k \in L_t \\t_k \neq t_i}} \exp( w_0 \cdot R_M^{(t_i , t_k)})
}
\end{equation}
where \( w_0 \in \mathbb{R}^+ \) is a temperature coefficient.

\subsubsection{Confusion Analysis Results}
\label{sec:3.2.5}
We integrate the implicit and explicit evaluations to obtain the final confusion analysis results for fine-grained entity types in the target domain:
\begin{equation}
\begin{split}
v_{conf}(t_i \rightarrow t_j) = v_{iml}(t_i \rightarrow t_j) + v_{exl}(t_i \rightarrow t_j)
\end{split}
\end{equation}
where \( v_{conf}(t_i \to t_j) \) is the model's confusion extent from \( t_i \) to \( t_j \). For each type \( t_i \), we select the entity type with the highest confusion extent as its most easily confused type:
\begin{equation}
T_{conf}^{(t_i)} = \underset{t_j \in L_t, t_j \neq t_i}{\arg\max} \left( v_{conf}(t_i \rightarrow t_j) \right)
\end{equation}
where \(T_{conf}^{(t_i)}\) is the most easily confused entity type for the entity type \( t_i \). MCQM provides a comprehensive analysis of the model's confusion extent between fine-grained entity types, playing a critical role in the bidirectional reasoning of PBRC.

\begin{figure}
    \centering
    \includegraphics[width=1\linewidth]{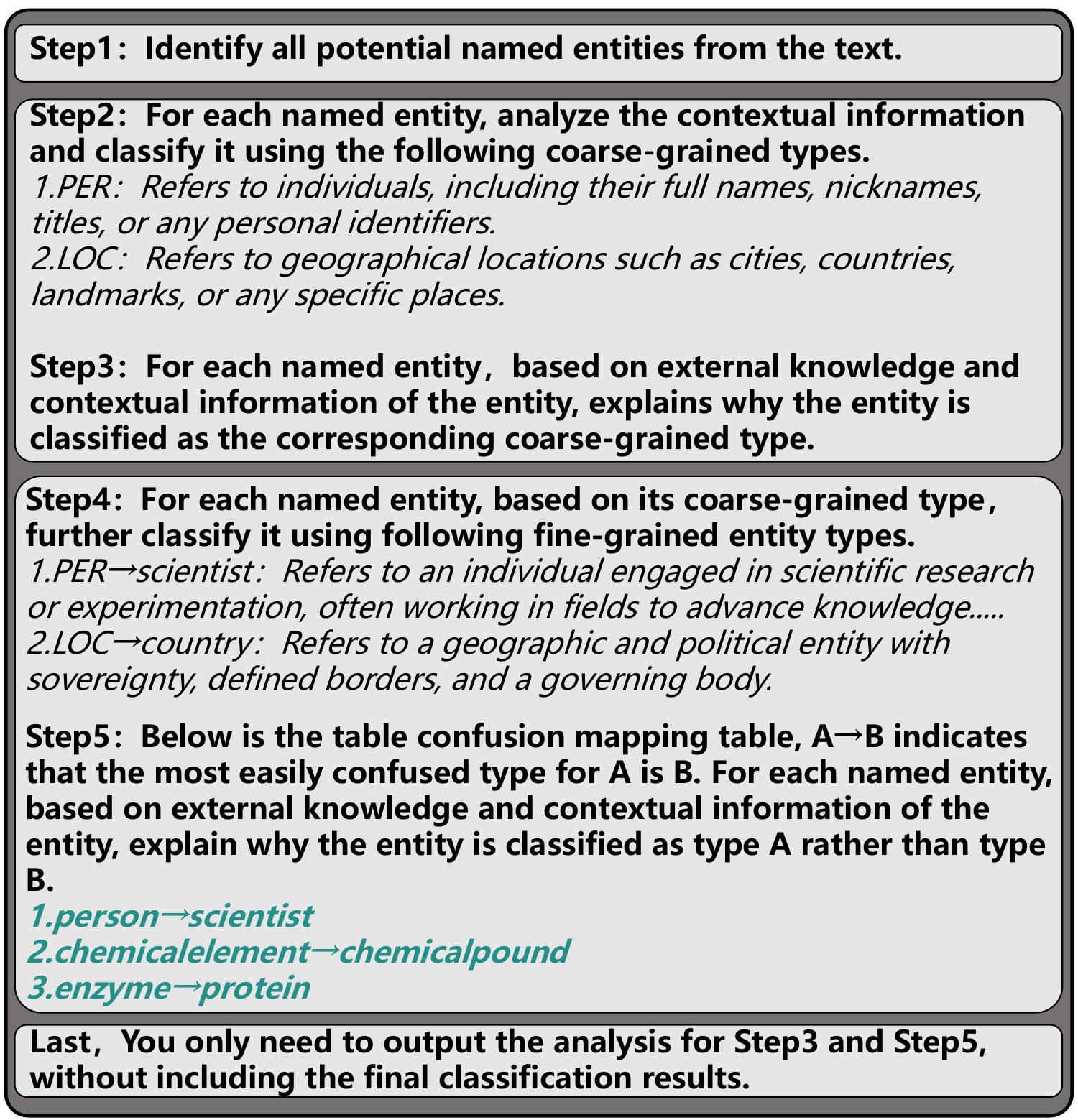}
    \caption{Contents of PBRC.}
    \label{fig3}
\end{figure}

\subsection{Progressive Bidirectional Reasoning Chain}
\label{sec:3.3}
\subsubsection{Contents of PBRC}
\label{sec:3.3.1}
We propose PBRC based on the source-target hierarchy and the confusion analysis from MCQM. As shown in Figure~\ref{fig:2}, PBRC prompts the LLM to output two-stage reasoning information. 
Figure~\ref{fig3} shows the contents of PBRC.

\textbf{Step1: Identifying Named Entities.} This step instructs the LLM to identify all potential named entities in the text. With pre-trained multi-domain knowledge and contextual understanding, the LLM is capable of recognizing potential entities in texts across various domains.

\textbf{Step2 and Step3: Coarse-grained Classification and Initial Reasoning.} We provide all source domain coarse-grained entity types and explanations. For each entity in Step1, we instruct the LLM to conduct coarse-grained classification and provide explanations, leveraging both its prior knowledge and an initial analysis of the context. This step unifies entity types across different domains, serving as an initial classification and reasoning process for entities.

\textbf{Step4 and Step5: Fine-grained Classification and Bidirectional Reasoning.} We provide all fine-grained entity types of the target domain along with their corresponding source domain prefixes and explanations. For each entity, we instruct the LLM to conduct fine-grained classification based on the initial classification and reasoning information from Step2 and Step3, and then conduct bidirectional reasoning through a deep analysis of contextual information and the LLM's rich external knowledge. As shown in Figure~\ref{fig3}, we present the confusion analysis result of the MCQM, where \(A\to B \) indicates that \( A \)'s most easily confused type is \( B \). Bidirectional reasoning consists of two parts: (1) \textbf{forward reasoning}, which infers that the entity type is \( A \); (2) \textbf{backward reasoning}, which infers that the entity type is not \( B \).

\textbf{Two-stage Reasoning Information.} 
The LLM only outputs the initial reasoning from Step3 and the bidirectional reasoning from Step5. 
The two-stage reasoning information is fed into the backbone for knowledge augmentation: the initial reasoning guides the backbone to leverage the knowledge from the source domain, strengthening knowledge transfer; the bidirectional reasoning guides the backbone to differentiate easily confused entity types, mitigating the entity type confusion. Besides, external knowledge mitigates the insufficiency of contextual information.

\begin{table}
\small
\centering
\renewcommand{\arraystretch}{0.5} 
\setcellgapes{1pt} 
\makegapedcells 
\begin{tabular}{cccccc}
\toprule
\textbf{Domain} & \textbf{Dataset} & \textbf{Type} 
& \textbf{Train} & \textbf{Dev} & \textbf{Test} \\
\midrule
\multirow{2}{*}{\textbf{Source}} 
& CoNLL2003 & 4 & 14987 & - & - \\
& Twitter   & 4 & 4290  & - & - \\
\midrule
\multirow{5}{*}{\textbf{Target}} 
& politics   & 9  & 200 & 541 & 651 \\
& science    & 17 & 200 & 450 & 543 \\
& music      & 13 & 100 & 380 & 465 \\
& literature & 12 & 100 & 400 & 416 \\
& AI         & 14 & 100 & 350 & 431 \\
\bottomrule
\end{tabular}
\caption{The statistics of all datasets.}
\label{tab:1}
\end{table}

\begin{table*}
\centering
\small
\setlength{\tabcolsep}{5pt}
\renewcommand{\arraystretch}{0.5} 
\setcellgapes{1pt}
\makegapedcells 
\begin{tabular}{l|cccccc|cccccc}
\toprule
 & \multicolumn{6}{c|}{\normalsize{\textbf{CoNLL2003}}} & \multicolumn{6}{c}{\normalsize{\textbf{Twitter}}} \\
\textbf{\normalsize Method} & \textbf{\footnotesize sci.} & \textbf{\footnotesize pol.} & \textbf{\footnotesize mus.} & \textbf{\footnotesize lit.} & \textbf{\footnotesize AI} & \textbf{\footnotesize Avg.} & \textbf{\footnotesize sci.} & \textbf{\footnotesize pol.} & \textbf{\footnotesize mus.} & \textbf{\footnotesize lit.} & \textbf{\footnotesize AI} & \textbf{\footnotesize Avg.} \\
\hline
MTD & 72.35 & 76.70& 76.10& 69.22 & 68.93 & 72.66 & 71.37 & 74.62 & 74.41 & 69.67 & 64.55 & 70.92 \\
CP-NER & 75.82 & 74.25 & 79.10 & 72.17 & 67.95 & 73.86 & - & - & - & - & - & - \\
DH-GAT & 74.21 & 77.06 & 78.77 & 72.51 & 69.30 & 74.37 & 74.55 & 76.46 & 77.33 & 71.52 & 67.65 & 73.50 \\
PromptNER & 72.59 & 78.61 & 84.26 & 74.44 & 64.83 & 74.95 & - & - & - & - & - & - \\
Dual-CL & 74.17 & 77.56 & 78.57 & 72.43 & 69.49 & 74.44 & 74.07 & 77.53 & 77.21 & 71.72 & 68.71 & 73.85 \\
DT-MPrompt & 73.06 & 80.54 & 79.54 & 73.51 & 70.13 & 75.36 & 73.18 & 79.86 & 77.93 & 72.74 & 69.13 & 74.57 \\
IF-WRANER-7B & 75.31 & 79.80 & 85.43 & 75.52 & 68.81 & 76.97 & - & - & - & - & - & - \\
TOPT & 80.16 & 81.55 & 82.03 & 77.85 & 72.34 & 78.78 & - & - & - & - & - & -\\
\hline
Flan-T5-base + MCQM \& PBRC & 81.43 & 83.53 & 83.69 & 79.03 & 74.29 & 80.39 & 80.32 & 82.43 & 81.93 & 77.62 & 73.97 & 79.25 \\
Flan-T5-large + MCQM \& PBRC & \textbf{82.34} & \textbf{85.03} & \textbf{85.62} & \textbf{79.96} & \textbf{75.99} & \textbf{81.79} & \textbf{81.62} & \textbf{84.64} & \textbf{84.02} & \textbf{78.61} & \textbf{75.50} & \textbf{80.88} \\
\bottomrule
\end{tabular}
\caption{F1 scores on CrossNER: CoNLL2003 and Twitter as source domains, respectively. Bold marks the highest.}
\label{tab:2}
\end{table*}

\begin{table*}
  \centering
  \setcellgapes{1pt} 
  \makegapedcells    
  \begin{tabular}{l|ccc|ccc}
    \toprule
     & \multicolumn{3}{c|}{\textbf{Flan-T5-base (w/o MCQM \& PBRC)}} & \multicolumn{3}{c}{\textbf{Flan-T5-base (+ MCQM \& PBRC)}} \\
    \cmidrule{2-4} \cmidrule{5-7}
                 \textbf{Domain}   & \textbf{Precision} & \textbf{Recall} & \textbf{F1 score} & \textbf{Precision} & \textbf{Recall} & \textbf{F1 score} \\
    \hline
    \normalsize{science}   &  71.31 \small{(83.11)} & 68.66 \small{(81.10)} & 69.96 \small{(82.09)}  & 83.05 \small{(88.12)}& 79.88 \small{(84.49)}  & 81.43 \small{(86.27)} \\
    
    \normalsize{politics}   &  71.30 \small{(87.45)}        & 69.29 \small{(86.99)}         & 70.28 \small{(87.22)}         & 84.62 \small{(91.15)}   & 82.47 \small{(88.01)}         & 83.53 \small{(89.55)} \\
    
    \normalsize{music}   &  77.07 \small{(88.46)}   & 73.65 \small{(82.99)}   & 75.32 \small{(85.64)}  & 85.34 \small{(89.45)}  & 82.10 \small{(87.85)}  & 83.69 \small{(88.64)}\\
    
    \normalsize{literature}   &  68.28 \small{(83.67)}   & 67.47 \small{(83.98)} & 67.87 \small{(83.82)}         & 80.48 \small{(87.67)}         & 77.64 \small{(84.12)}        & 79.03 \small{(85.86)} \\
    
    \normalsize{AI}   &  63.42 \small{(81.15)}        & 63.70 \small{(81.64)}          & 63.56 \small{(81.39)}         & 75.74 \small{(86.07)}         & 72.90 \small{(83.01)}         & 74.29 \small{(84.51)} \\
    
    \bottomrule
  \end{tabular}
  \caption{Entity Type Confusion Analysis. CoNLL2003 as the source domain.}
  \label{tab:7}
\end{table*}

\subsubsection{Knowledge Augmentation}
\label{sec:3.3.2}
The two-stage reasoning information generated by the LLM is utilized to augment the knowledge of $\mathcal{M}_{src}$. Through supervised fine-tuning, the backbone learns to leverage this reasoning information, yielding the final backbone model $\mathcal{M}_{tgt}^\dagger$.
\begin{itemize}
\setlength{\leftmargin}{0pt}
\item \textbf{Input:} Find all entities of types \{politician, person, country ...\} in \{text\}.
\\Reasoning information: \{...\}.
\\Output format: \{"type 1": ["entity 1", "entity 2"], "type 2": ["entity 3"], ...\}.
\item \textbf{Gold Sequence:} \{"country": ["Afghanistan"], ”politician": ["Barack Obama"], ...\}.
\end{itemize}

\section{Experiments}
\subsection{Experimental Setup}
\label{sec:4.1}
As shown in Figure~\ref{tab:1}, we evaluate on two source domains (CoNLL2003~\cite{44} and Twitter~\cite{46}) and five target domains (CrossNER~\cite{24}: politics, science, music, literature, and AI). We employ Flan-T5~\cite{22} as the backbone model and GPT-4o-mini as the LLM.
To evaluate the performance of our proposed method, we compare it with the following baselines: (1) \textbf{MTD}~\cite{49}: A modular learning-based method that decomposes NER into span detection and type classification. (2) \textbf{CP-NER}~\cite{13}: Transforms NER as text-to-text task and introduces collaborative domain-prefix tuning based T5 as well. (3) \textbf{DH-GAT}~\cite{11}: Applies Graph Attention Networks to encode syntactic and semantic information while embedding words into hyperbolic space. (4) \textbf{PromptNER}~\cite{41}: Uses GPT4 for NER through prompt templates. (5) \textbf{Dual-CL}~\cite{27}: Uses dual contrastive learning to refine ambiguous representations and learn generalizable features. (6) \textbf{DT-MPrompt}~\cite{36}: Splits the cross-domain NER task into subtasks and uses separate functional modules for learning and knowledge transfer. (7) \textbf{IF-WRANER}~\cite{48}: Fine-tunes 7B LLaMA with instruction fine-tuning and employs word embeddings to retrieve examples for in-context learning. (8) \textbf{TOPT (SOTA)}~\cite{53}: Utilizes LLaMA to generate task-oriented knowledge for flan-T5 and adopts task-oriented pre-training for domain adaptation.

\subsection{Main Results}
\label{sec:4.2}
The main results are shown in Table~\ref{tab:2}.

\textbf{CoNLL2003 as the Source Domain.} It is observed that the F1 score improves as the parameter scale of the model increases. On average, our method achieves consistent F1 score improvements of +1.61\% (Flan-T5-base) and +3.01\% (Flan-T5-large). Our method surpasses prior SOTA results in science, politics, literature, and AI domains (effective for both Flan-T5-base and Flan-T5-large), achieving improvements exceeding 2.10\% in these domains. These results demonstrate the strong effectiveness of two-stage information generated by the LLM in cross-domain NER. We observe that ~\textbf{IF-WARNER-7B} (85.43\%), which employs a fine-tuned 7B LLaMA, and \textbf{PromptNER} (84.26\%), which utilizes GPT-4, both achieve significantly higher F1 scores in the music domain compared to other baselines using smaller models. This demonstrates the advantage of LLMs in the music domain, ensuring the quality of the information generated by our two-stage process using the LLM. We observe that Flan-T5-large achieves only a marginal improvement of 0.19\% over prior SOTA in the music domain, while Flan-T5-base fails to surpass the prior SOTA. Through careful analysis, we have identified the main reasons. First, the high overlap between the domain knowledge of the LLM and Flan-T5 in the music domain diminished their complementary effects, resulting in limited benefits from two-stage reasoning. Then, though the two-stage information significantly improves F1 scores for certain entity types, their limited entity count results in a negligible impact on the overall F1 metric.

\textbf{Twitter as the Source Domain.} Our method consistently achieves significant improvements across all domains.
Since \textbf{TOPT} is not tested with Twitter as the source domain, we cannot compare with it directly. Compared to CoNLL2003, the performance of both Flan-T5-base and Flan-T5-large shows a slight decline. We attribute this to the smaller dataset size of Twitter, which limits the model's ability to learn coarse-grained features of entities. This highlights the importance of learning the coarse-grained features of entities and further demonstrates that our method enables effective knowledge transfer from the source to the target domain.

\begin{table}
  \centering
  \begin{tabular}{l ccccc}
    \toprule
    \normalsize \textbf{Setting} & \normalsize \textbf{sci.} & \normalsize \textbf{pol.} & \normalsize \textbf{mus.} & \normalsize \textbf{lit.} & \normalsize \textbf{AI} \\
    \midrule
    w/o   & 78.06   & 79.47   & 81.44   & 76.12   & 71.42   \\				
    + SS &   79.01     &  82.62   & 82.54      &    77.34        &  73.05  \\
    + STH &   79.98     &  82.67       & 82.09      &   77.86         & 73.59   \\
    + SS\&STH &  80.93      & 83.05  &  83.16     & 78.41     &   73.66   \\
    + DE &  79.47      & 82.89        &  82.41     &   77.49         & 72.18   \\
    \hline
    + ALL & \textbf{81.43}  & \textbf{ 83.53} &  \textbf {83.69} & \textbf {79.03} &  \textbf{74.29}  \\
    \bottomrule
  \end{tabular}
  \caption{Ablation Study on MCQM (F1 scores).}
  \label{tab:8}
\end{table}

\subsection{Entity Type Confusion Analysis}
To verify that the model’s performance improvement is attributed to enhanced entity type classification capabilities, we evaluate its performance in both entity span detection and overall performance using precision, recall, and F1 score. 
The results are shown in Table~\ref{tab:7}, where the values in parentheses indicate the performance on entity span detection. 
Before incorporating MCQM and PBRC, we observe a substantial performance gap between the entity span detection and overall performance, which indicates that the model’s ability of entity type classification is the primary bottleneck, suffering from significant entity type confusion. 
After incorporating MCQM and PBRC, we observe that our method improves the F1 score for entity span detection by 2\%–4\% across the five domains, with slight improvements also observed in precision and recall. More notably, the overall performance increases by over 10\% on average. These results demonstrate that our method significantly enhances the model’s ability in entity type classification and effectively mitigates the model's entity type confusion.

\subsection{Ablation Study}
\label{4.3}
\subsubsection{Ablation Study on MCQM}

Table~\ref{tab:8} presents the effects of MCQM components: semantic similarity (SS), source–target hierarchy (STH), and data evaluation (DE), on model performance (F1 score). 
By comparing the results of completely removing all components (w/o) with those of introducing only SS, STH, or DE, we observe that, across all target domains, incorporating SS, STH, or DE into confusion analysis and subsequently introducing backward reasoning based on the corresponding analysis improves model performance. This indicates that semantic similarity analysis, source–target hierarchy analysis, and explicit data evaluation can all effectively quantify the entity type confusion extent.
Moreover, simultaneously introducing SS and STH (+SS\&STH) or incorporating all components (+ALL) into confusion analysis further improves model performance, suggesting that leveraging multiple analytical dimensions can reduce the errors associated with a single dimension and enhance the quantification accuracy of MCQM.

\subsubsection{Ablation Study on PBRC}

\begin{table}
\centering
\setlength{\tabcolsep}{4.5pt}
\setcellgapes{2.5pt} 
\makegapedcells 
\begin{tabular}{lcccccc}
\toprule
\textbf{Setting}& \textbf{sci.}& \textbf{pol.}& \textbf{mus.}& \textbf{lit.}& \textbf{AI}& \textbf{Avg.}
\\
\hline
w/o & 69.96 & 70.28 & 75.32 & 67.87 & 63.56 & 69.40
\\
+ FR & 76.58 & 78.26 & 80.09 & 74.77 & 70.03 & 75.95
\\
+ IR\&FR & 78.06 & 79.47 & 81.44 & 76.12 & 71.42 & 77.30
\\
+ BR\&FR & 80.62 & 82.76 & 82.49 & 78.06 & 73.52 & 79.49
\\
\hline
+ ALL & \textbf{81.43} & \textbf{83.53} & \textbf{83.69} & \textbf{79.03} & \textbf{74.29} & \textbf{80.39}
\\
\bottomrule
\end{tabular}
\caption{Ablation Study on PBRC (F1 scores).}
\label{tab:3}
\end{table}

We conduct ablation studies to evaluate the effectiveness of each reasoning component in PBRC: forward (FR), backward (BR), and initial reasoning (IR). As shown in Table~\ref{tab:3}, adding forward reasoning (FR) alone yields an average performance gain of 6.55\%. This demonstrates that the reasoning information provided by the LLM can effectively support the backbone model in performing NER.
Further incorporating initial reasoning (IR) brings an additional average improvement of 1.35\%, demonstrating that the two-stage reasoning from source to target domain can enhance the backbone’s cross-domain transferability.
Moreover, when backward reasoning (BR) is incorporated, the average performance rises by 3.54\% compared to using FR alone. This indicates that the MCQM effectively analyzes and quantifies the model's confusion among entity types, while backward reasoning can alleviate entity type confusion.

\subsection{Runtime Analysis}
\begin{table}
\centering
\begin{tabular}{l ccc}
\toprule
\textbf{Model} & \textbf{\small {Training}} & \textbf{\small {Inference}} & \textbf{\small {Dev F1}} \\
\midrule
GPT-4o-mini  & - & {3.22s} & {74.78} \\
Flan-T5-base  & {28.62 min}  & {0.32s} & {70.25} \\
GPT-4o-mini + base  & {23.60 min} & {3.56s} & {80.69} \\
Flan-T5-large  & {46.28 min} & {0.51s} & {71.98} \\
GPT-4o-mini + large & {37.35 min} & {3.84s} & {82.55} \\
\bottomrule
\end{tabular}
\caption{The runtime statistics of different methods.}
\label{table:20}
\end{table}

We report the average training time of different methods across the five target domains, the average F1 score on the dev set, and the inference time per text (in seconds), as shown in Table~\ref{table:20}. First, although our method requires the LLM to generate reasoning information for the training set in advance, the overall training time is still lower than that of using Flan-T5 alone. We attribute this to the fact that, once reasoning information is introduced, Flan-T5 converges faster and reaches the best F1 score of the dev set with fewer epochs. Besides, for the inference time of each text, our method first leverages the LLM to generate reasoning information before feeding it into Flan-T5 for NER. Since Flan-T5 has a very short inference time, the final inference time of our method is comparable to that of the LLM. Our method maintains training time comparable to that of using Flan-T5 alone, attains inference efficiency that is nearly equivalent to the LLM, and achieves performance that substantially exceeds both Flan-T5 and the LLM when employed alone. 

\section{Conclusion}
In this paper, we propose the MCQM, which effectively quantifies the model's confusion extent among fine-grained entity types from three dimensions: source-target hierarchy analysis, semantic similarity analysis, and explicit data evaluation. Furthermore, we propose the PBRC based on the source-target hierarchy and the MCQM, which can significantly mitigate entity type confusion and improve the model's generalization ability. Our method achieves SOTA results on all domains of the CrossNER dataset.

\section*{Acknowledgments}
This work is partially supported by the National Natural Science Foundation of China (No. 62272045).

\bibliographystyle{named}
\bibliography{ijcai26}

\end{document}